\documentclass[10pt,twocolumn,letterpaper]{article}

\usepackage{iccv}
\usepackage{times}
\usepackage{epsfig}
\usepackage{graphicx}
\usepackage{amsmath}
\usepackage{amssymb}

\usepackage[breaklinks=true,bookmarks=false]{hyperref}

\iccvfinalcopy 

\def\iccvPaperID{12} 

\ificcvfinal\fi

\begin{document}
\title{Video Generation with Consistency Tuning}

\author{Chaoyi Wang, Yaozhe Song, Yafeng Zhang, Jun Pei, Lijie Xia, Jianpo Liu\\
	Chinese Academy of Sciences, Shanghai Institute of Microsystem and Information Technology\\
	{\tt\small chaoyiwang, YaozheSong, acampus, peijun, xialj, liujp@mail.sim.ac.cn}
}
\maketitle
\ificcvfinal\thispagestyle{empty}\fi

\begin{abstract}

  Currently, various studies have been exploring generation of long videos. However, the generated frames in these videos often exhibit jitter and noise. Therefore, in order to generate the videos without these noise, we propose a novel framework composed of four modules: separate tuning module, average fusion module, combined tuning module, and inter-frame consistency module. By applying our newly proposed modules subsequently, the consistency of the background and foreground in each video frames is optimized. Besides, the experimental results demonstrate that videos generated by our method exhibit a high quality in comparison of the state-of-the-art methods~\cite{wang2023gen,ceylan2023pix2video,wu2022tune}.
\end{abstract}

\section{Introduction}

Recently, diffusion models have achieved great success in handling complex and large-scale image datasets~\cite{dhariwal2021diffusion}, these methods have shown great potential to model video distribution much better with scalability both in terms of spatial resolution and temporal duration~\cite{ho2022imagen}. Besides, a current method extends off-the-shelf short video diffusion models for generating long video~\cite{wang2023gen}.

Despite advances in video generation, the generated videos still present certain shortcomings. For example, unnatural jumps always occur between frames of generated videos. Therefore, we propose an automated video processing framework to address the aforementioned shortcomings, which includes four modules: separate diffusion module, average fusion module, combined tuning module and inter-frame consistency module.
\section{Related Work}
In recent years, the generation of long videos based on text has drawn tremendous attention, leading to various attempts to tackle the challenges associated with this task~\cite{voleti2022mcvd,yu2023video,ge2022long}. Denoising Diffusion Probabilistic Model (DDPM)~\cite{ho2020denoising} and its variant Denoising Diffusion Implicit Model (DDIM)~\cite{song2020denoising} have been widely used for text-to-image generation, including MCVD~\cite{voleti2022mcvd}, FDM~\cite{harvey2022flexible},
 LVDM~\cite{he2022latent}, PVDM~\cite{yu2023video}, Gen-L-Video~\cite{wang2023gen} and so on. In the presence of high quality text conditioned image generation models, several recent works have focused on utilizing additional control signals for generation or editing existing images~\cite{wang2023gen}. However, these methods tend to have some limitations for practical applications that maintaining consistency is challenging as frame count increases in the video generation. 

Therefore, in order to generate videos with consistency between frames, We demonstrate the above four modules in our proposed framework. 

\section{Methods}
\begin{figure}[h]
	\begin{center}
		\includegraphics[width=0.95\linewidth]{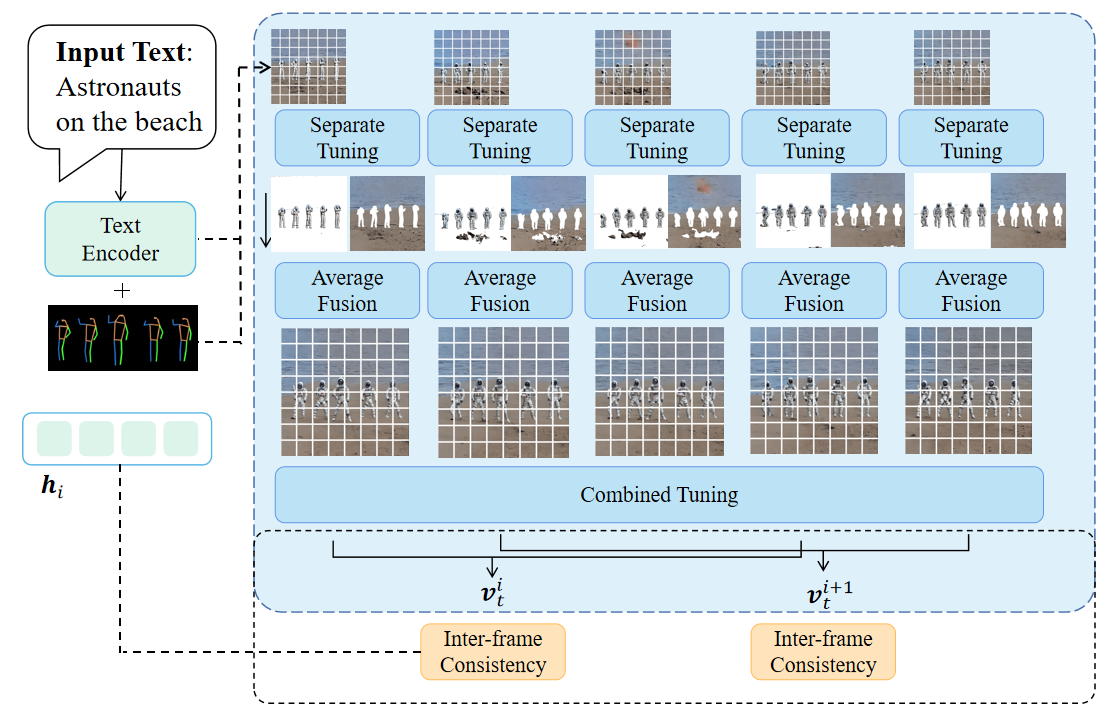}
	\end{center}
	\caption{A high level overview of the proposed video generation approach. }
	\label{fig:overview}
\end{figure}

%
We develop a framework to generate video with high quality context and smooth frames composed of four modules in sequence, separate tuning module, average fusion module, combined tuning module and inter-frame consistency module. The detailed implementation of the methods are shown in figure~\ref{fig:overview}.

\subsection{Separate Tuning Module}
Given the input prompts of the hidden embeddings $\boldsymbol{h}$, and the corresponding condition $\boldsymbol{c}$, image frames $\boldsymbol{r}$ are created by 
\begin{align}
    \boldsymbol{r} = \mathcal{F}_{0}(\boldsymbol{h},\boldsymbol{c})
\end{align}
where $\mathcal{F}_{0}$ denotes the function of the diffusion and denosing process. 

In details, in the whole diffusion and denoising process, the diffusion models~\cite{sohl2015deep} are trained through latent variable representation of the form $p_{\theta}(\boldsymbol{x}):=\int p_{\theta}(\boldsymbol{x}_{0:T})d_{\boldsymbol{x}_{1:T}}$ where $p_{\theta}(\boldsymbol{x}_{0:T})$ is the joint distribution and $\boldsymbol{x}_{1},...,\boldsymbol{x}_{T}$ are latents. The data are perturbed by gradually introducing noise to $\boldsymbol{x}_0 \sim q(x_0)$, formalized by a Markov chain, known as forward process:
\begin{align}
	q(\boldsymbol{x}_{1:T}|\boldsymbol{x}_{0}) &= \prod\limits_{t=1}^{T} q(\boldsymbol{x}_{t}|\boldsymbol{x}_{t-1})\\
	q(\boldsymbol{x}_{t}|\boldsymbol{x}_{t-1}) &= \mathcal{N}(\boldsymbol{x}_{t}|\sqrt{\alpha_{t}}\boldsymbol{x}_{t-1},\beta_{t}\mathbf{I})
\end{align}
where $\beta_{t}$ is the noise schedule and $\alpha_{t} = 1-\beta_{t}$. The diffusion model $p_{\theta}(\boldsymbol{x}_{t=1}|x_{t})$ parameterized by $\theta$ is trained to approximate the reverse transition $q(\boldsymbol{x}_{t-1}|\boldsymbol{x}_t,\boldsymbol{x}_0)$, which is formulated as 
\begin{align}
	q(\boldsymbol{x}_{t-1}|\boldsymbol{x}_{t},\boldsymbol{x}_{0}) = \mathcal{N}(\boldsymbol{x}_{t-1};\tilde{\boldsymbol{\mu}}(\boldsymbol{x}_{t},\boldsymbol{x}_{0}),\tilde{\beta}_{t}\mathbf{I})\\
	\tilde{\boldsymbol{\mu}}(\boldsymbol{x}_{t},\boldsymbol{x}_{0}) := \frac{\sqrt{\bar{\alpha}_{t-1}}\beta_{t}}{1-\bar{\alpha}_{t}}\boldsymbol{x}_{0} + \frac{\sqrt{\alpha_{t}}(1-\bar{\alpha}_{t-1})}{1-\bar{\alpha}_{t}}\boldsymbol{x}_{t}
\end{align}
where $\bar{\alpha}_{t}:=\prod_{s=1}^{t}\alpha_{s}$ and $\tilde{\beta}_{t}:= \frac{1-\bar{\alpha}_{t-1}}{1-\bar{\alpha}_{t}}\beta_{t}$.

The loss function of a noise prediction network is denoted as $\boldsymbol{\epsilon}_{\theta}(\boldsymbol{x}_{t},t)$:
\begin{align}
	\mathcal{L}_{1}(\theta) := \mathbb{E}_{t,\boldsymbol{x}_{0},\boldsymbol{\epsilon}} \lVert \boldsymbol{\epsilon} - \boldsymbol{\epsilon}_{\theta}(\sqrt{\bar{\alpha}_{t}}\boldsymbol{x}_{0}+\sqrt{1-\bar{\alpha}} \boldsymbol{\epsilon},t) \rVert^2
\end{align}
where $t$ is uniform between $1$ and $T$.

Then, denoising diffusion implicit model is learned to (DDIM)~\cite{song2020denoising} generalize the framework of denoising diffusion probabilistic model (DDPM)~\cite{ho2020denoising} and propose a deterministic ODE process, achieving faster sampling speed.

At each diffusion step, given a noisy sample $\boldsymbol{x}_t$, a prediction of the noise-free sample $\hat{\boldsymbol{x}}_{0}$ along with a direction that points to $\boldsymbol{x}_t$ is computed. The final prediction of $\boldsymbol{x}_{t-1}$ is obtained by:
\begin{align}
	\boldsymbol{x}_{t-1} &=\sqrt{\alpha_{t-1}} \hat{\boldsymbol{x}}_{0}^{t} + \sqrt{1-\alpha{_{t-1}}-\sigma_{t}^{2}}\epsilon_{\theta}(\boldsymbol{x}_{t},t) + \sigma_{t}\epsilon_{t}\\
	\hat{\boldsymbol{x}}_{0}^{t}&=\frac{\boldsymbol{x}_{t}-\sqrt{1-\alpha_{t}}\epsilon_{\theta}(\boldsymbol{x}_{t})}{\sqrt{\alpha_{t}}}
\end{align}
where $\alpha_{t}$ and $\sigma_{t}$ are the parameters of the scheduler and $\epsilon_{\theta}$ is the noise predicted by the decoder at the current step $t$. 

Therefore, the denoised image frames $\boldsymbol{r}$ is denoted as $\hat{\boldsymbol{x}}_{0}^{t}$.

To get the corresponding background and foreground, 
\begin{align}
    [\boldsymbol{m}_{fg} ,\boldsymbol{m}_{bg}]=\mathcal{F}_{1}(\boldsymbol{r})
\end{align}
where $\mathcal{F}_{1}$ is implemented by employ Segment Anything Model (SAM)~\cite{kirillov2023segment} to get $\boldsymbol{m}_{fg}$ (foreground) and $\boldsymbol{m}_{bg}$ (background). 

\subsection{Average Fusion Module}
\begin{figure}[h]
	\begin{center}
		\includegraphics[width=1 \linewidth]{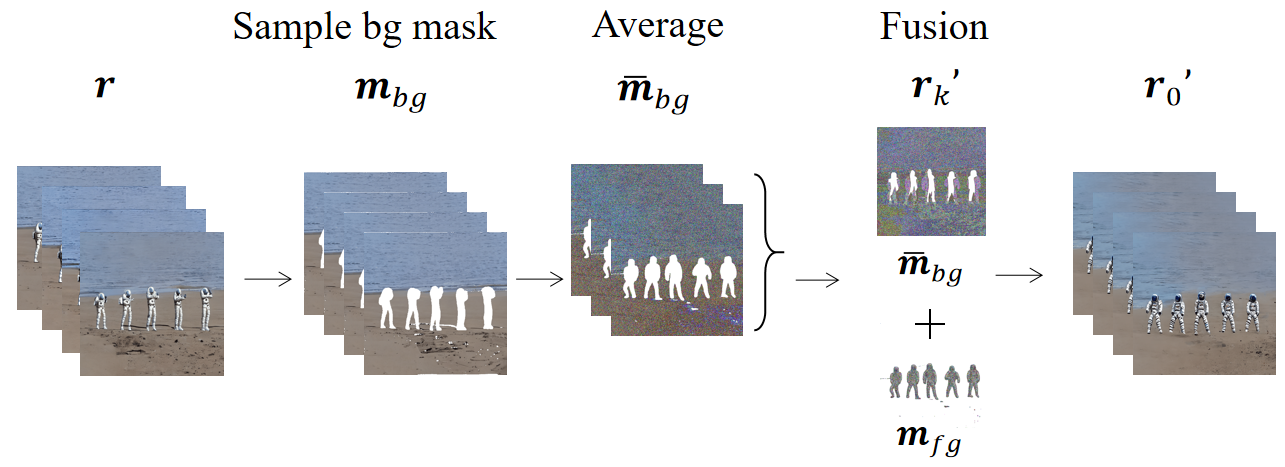}
	\end{center}
	\caption{The flow diagram of the proposed average fusion method.}
	\label{fig:average}
\end{figure}

The key objective of this module is to optimize the consistency of the image frames to make them have a smooth transition between frames and maintain the foreground area for future fine-tuning. Given the original image frames $\boldsymbol{r}$ and their masked areas $\boldsymbol{m}_{fg}$,  $\boldsymbol{m}_{bg}$, we can obtain an optimal $\boldsymbol{r}_{k}'$ at certain step $t = k$ in the diffusion model:
\begin{align}
    \boldsymbol{r}_{k}' = \mathcal{F}_{2}(\boldsymbol{r},\boldsymbol{m}_{fg},\boldsymbol{m}_{bg})
\end{align}
$\mathcal{F}_{2}$ is denoted by:
\begin{align}
	\boldsymbol{r}_{k}' = \frac{1}{n}\cdot w_{1}\sum\limits_{i=1}^{n}\boldsymbol{m}_{{bg}_{i,k}} + w_{2}\cdot \boldsymbol{m}_{{fg}_k}
\end{align}
where $\boldsymbol{m}_{{bg}_{i,k}}$ is the $i$th masked background areas randomly sampled from $\boldsymbol{r}$ at step $k$; $\boldsymbol{m}_{{fg}_k}$ is the masked foreground area of $\boldsymbol{r}_{k}$ at step $k$, $w_1, w_2$ are the learnable weights for background and foreground. The flow diagram of the procedures is illustrated in figure~\ref{fig:average}.

Then we conduct the remaining denoising procedures via equation (8) to get $\boldsymbol{r}_{0}'$ denoting as  $\boldsymbol{r}'$:
\begin{align}
    \boldsymbol{r}' = \mathcal{F}_{3}(\boldsymbol{r}_{k}', \boldsymbol{h},\boldsymbol{c})
\end{align}
where $\mathcal{F}_{3}$ is the function of the denoising model with modified $\boldsymbol{r}_{k}'$ at step $k$.

\subsection{Combined Tuning Module}
Afterward, a proposed fine-tune process to optimize the diffusion model parameters with $\boldsymbol{r}'$ and text embeddings $\boldsymbol{h}$ is conducted. This is achieved by the following equations:

\begin{align}
   \mathcal{D}'  = \mathcal{F}_4(\mathcal{D}, \boldsymbol{r}', \boldsymbol{h})
\end{align}
where $\mathcal{D}$ is the diffusion model with a denoising model $\boldsymbol{\epsilon}_{\boldsymbol{\theta}}$ of parameter $\boldsymbol{\theta}$.

This procedure  $\mathcal{F}_{4}$ can be formulated as an optimization problem:
\begin{align}
\arg\min\limits_{\boldsymbol{\theta}}\mathbb{E}_{\boldsymbol{x}_{0},\boldsymbol{x}_{t},\boldsymbol{\epsilon}\sim\mathcal{N}(0,\mathbf{I})}\lVert  (\boldsymbol{\epsilon}-\boldsymbol{\epsilon}_{\theta}(\boldsymbol{x}_{t},t,\boldsymbol{h})) \rVert^2
\end{align}

To better utilize the properties of foreground and background, We apply separate weights to the foreground and background areas to finely tune the output video effects.
\begin{align}
	&\mathcal{L}_{fg}= {E}_{\boldsymbol{x}_{0},\boldsymbol{x}_{t},\boldsymbol{\epsilon}\sim\mathcal{N}(0,\mathbf{I})}\lVert M_{fg}\otimes (\boldsymbol{\epsilon}-\boldsymbol{\epsilon}_{\theta}(\boldsymbol{x}_{t},t,\boldsymbol{h})) \rVert^2\\
	&\mathcal{L}_{bg} = {E}_{\boldsymbol{x}_{0},\boldsymbol{x}_{t},\boldsymbol{\epsilon}\sim\mathcal{N}(0,\mathbf{I})}\lVert M_{bg}\otimes (\boldsymbol{\epsilon}-\boldsymbol{\epsilon}_{\theta}(\boldsymbol{x}_{t},t,\boldsymbol{h})) \rVert^2
\end{align}
where $\otimes$ represent the Hadamard product, $\mathcal{L}_{fg}$ is the loss of foreground, $\mathcal{L}_{bg}$ is the loss of background,  $M_{fg}$ and $M_{bg}$ are masks of foreground and background.
\subsection{Inter-frame Consistency Module}
Afterwards, the inter-frame consistency module~\cite{wang2023gen} is proposed as follows:
\begin{align}
    \boldsymbol{v} = \mathcal{F}_{5}(\mathcal{D}',\boldsymbol{h},\boldsymbol{c})
\end{align}

$\mathcal{F}_{5}$ considers the denoising process of the entire video as multiple short videos with
temporal overlapping undergoing parallel denoising in the temporal domain \cite{wang2023gen}. Suppose a model $\mathcal{D}_{\boldsymbol{\theta}}^{l}(\boldsymbol{v}_{t-1}|\boldsymbol{v}_{t},\boldsymbol{h})$ in layer $l$ with corresponding noise prediction network $\boldsymbol{\epsilon}_{\boldsymbol{\theta}}^{l}(\boldsymbol{v}_{t},t,\boldsymbol{h})$ capable of denoising the given long video $\boldsymbol{v}_{t}$ such that $\boldsymbol{v}_{t-1}\sim \mathcal{D}_{\boldsymbol{\theta}}^{l}(\boldsymbol{v}_{t-1}|\boldsymbol{v}_{t},\boldsymbol{h})$.

Define a mappings $\mathcal{P}_i$ which projects original videos $\boldsymbol{v}_{t}$ in the trajectory to short video segments $\boldsymbol{v}_{t}^{i}$:
\begin{align}
    \mathcal{P}_{i}(\boldsymbol{v}_{t})=\boldsymbol{v}_{t}^{i} = \boldsymbol{v}_{t,S\times i:S\times i+ K}
\end{align}
where $t=1,2,...,T$, $i = 0,1,...,N-1$, $\boldsymbol{v}_{t,S\times i:S\times i+ K}$ represents the collection of frames with sequence number from $S\times i$ to $S \times i + K$, S represents the stride among adjacent short video clips, $K$ is the fixed length of short videos, $N$ is the total number of clips. Each short video $\boldsymbol{v}_{t}^{i}$ is guided with an independent text condition $\boldsymbol{h}^{i}$. 

For simplicity, the diffusion models for all video clips are set to one single model $\mathcal{D}_{\boldsymbol{\theta}}'(\boldsymbol{v}_{t-1}^i|\boldsymbol{v}_{t}^i,\boldsymbol{h}^i)$.

The optimal $\boldsymbol{v}_{t-1}$ can finally be obtained by solving the following optimization problem:
\begin{align}
    \boldsymbol{v}_{t-1}=\arg\min\limits_{\boldsymbol{v}}\sum\limits_{i=0}^{N-1}\lVert W_{i}\otimes (\mathcal{P}_{i}(\boldsymbol{v})-\boldsymbol{v}_{t-1}^{i})\rVert^{2}
\end{align}
where $W_i$ is the pixel-wise weight for the video clip $\boldsymbol{v}_{t}^{i}$, $\otimes$ is the Hadamard product. For an arbitrary frame $j$ in the video $\boldsymbol{v}_{t-1}$, it is equal to the weighted sum of all the corresponding frames in short videos that contain the $j$ frame~\cite{wang2023gen}.

\subsection{Training objective}
Therefore, the total loss $\mathcal{L}_{total}$ for the optimization problem of generating the videos is defined as follows:
\begin{align}
    \mathcal{L}_{total} = \lambda_{1}\cdot\mathcal{L}_{1} + \lambda_{2}\cdot\mathcal{L}_{fg} + \lambda_{3}\cdot\mathcal{L}_{bg}
\end{align}where $\mathcal{L}_{1},\mathcal{L}_{fg},\mathcal{L}_{bg}$ are loss functions derived from equation (6),(15),(16) and $\lambda_{1},\lambda_{2},\lambda_{3}$ are their corresponding weights.
\section{Experiments}
Our process involves editing of videos generated based on text.  The implementation details of the proposed framework for  video generation is shown as follows:
\begin{list}{\labelitemi}{\leftmargin=1em}
\setlength{\topmargin}{0pt}
\setlength{\itemsep}{0em}
\setlength{\parskip}{0pt}
\setlength{\parsep}{0pt}
\item Input the text prompt and conditional pose frames to generate the corresponding embedding, then load the stable diffusion model v1-4~\cite{rombach2022high} and T2I-Adapter~\cite{mou2023t2i} model for image generation. The $\beta$ in the DDPM noise scheduler starts at $8.5$e-4 and ends at $1.2$e-2, with the scaled linear method and total diffusion steps of $1000$.
\item Randomly sample $5$ results at $5$th step from $\boldsymbol{r}_{995}$ at the back process of DDPM and get $\boldsymbol{r}_{0}$, apply SAM to $\boldsymbol{r}_{0}$ to obtain the foreground $\boldsymbol{m}_{fg}$ and background $\boldsymbol{m}_{bg}$, create $\boldsymbol{r}_{995}'$ by equation (11) and estimate $\boldsymbol{r}_{0}'$ by $\boldsymbol{r}_{995}'$ via back process.
\item Fine tune the diffusion model with learning rate of $2$e-6 for $250$ steps and batch size $1$ with a modified loss function by equation (20) and output $\mathcal{D}'$  in equation (13). 
\item Process the inter-frame consistency module with the input of a modified diffusion model $\mathcal{D}'$, text embeddings $\boldsymbol{h}$, condition $\boldsymbol{c}$ and output the final long video $\boldsymbol{v}$ (see equation (17) ).
\end{list}

\section{Result}
\begin{figure*}
\begin{center}
\includegraphics[width=0.95\linewidth]{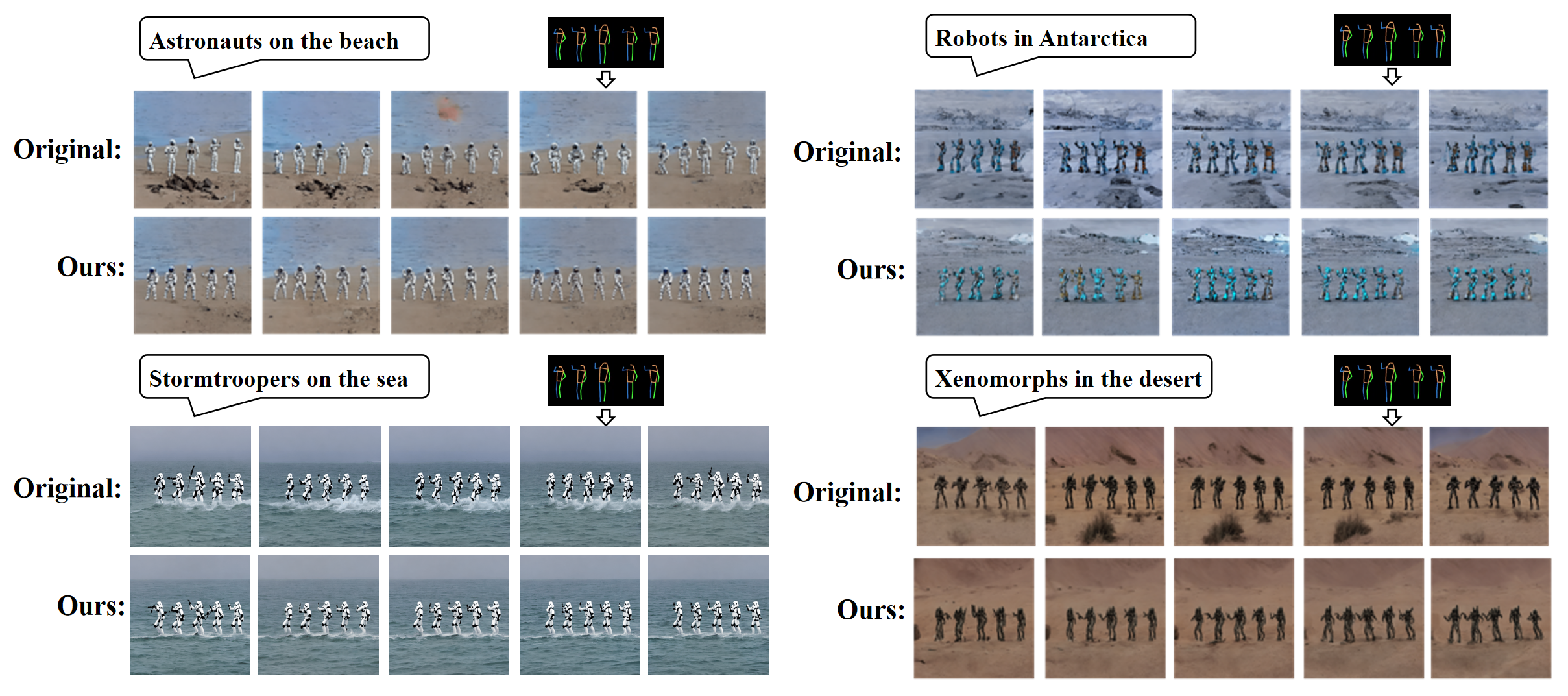}
\end{center}
   \caption{Comparison of frames between the state-of-the-art and our proposed framework.}
\label{fig:result}
\end{figure*}

Currently, we apply the first and second module (separate tuning module, average fusion module) and produce the initial result. We provide a visual presentation of our results in Figure~\ref{fig:result},  in comparison with the state-of-the-art~\cite{wang2023gen} and our proposed framework. It can be seen our method generates videos with a good consistency. We are going to apply the left modules and produce the final results.

\section{Conclusion}
We propose a video generation framework with four modules to generate a long video with good consistency. We complete our first experiment applying the separate tuning module and average fusion module, showing the experiment in comparison with the state-of-the-art. Then, we are going to apply the remaining modules and complete our left experiments.
{\small
\bibliographystyle{ieee_fullname}
\bibliography{egbib}
}

\end{document}